\documentclass[letterpaper]{article} 
\usepackage{aaai24}  
\usepackage{times}  
\usepackage{helvet}  
\usepackage{courier}  
\usepackage[hyphens]{url}  
\usepackage{graphicx} 
\usepackage{natbib}  
\usepackage{caption} 
\usepackage{algorithm}
\usepackage{algorithmic}

\usepackage{newfloat}
\usepackage{listings}
\DeclareCaptionStyle{ruled}{labelfont=normalfont,labelsep=colon,strut=off} 
\floatstyle{ruled}
\newfloat{listing}{tb}{lst}{}
\floatname{listing}{Listing}
\usepackage{amsmath}
\usepackage{amssymb}
\usepackage{arydshln}

\title{PNeSM: Arbitrary 3D Scene Stylization via Prompt-Based Neural Style Mapping}
\author{
    Jiafu Chen\textsuperscript{\rm 1}, Wei Xing\textsuperscript{\rm 1}$^*$, Jiakai Sun\textsuperscript{\rm 1}, Tianyi Chu\textsuperscript{\rm 1}, Yiling Huang\textsuperscript{\rm 1}\\
    Boyan Ji\textsuperscript{\rm 1}, Lei Zhao\textsuperscript{\rm 1}\thanks{Corresponding Authors.}, Huaizhong Lin\textsuperscript{\rm 1}$^*$, Haibo Chen\textsuperscript{\rm 2}, Zhizhong Wang\textsuperscript{\rm 1}$^*$
}
\affiliations{
    \textsuperscript{\rm 1}Zhejiang University\\
    \textsuperscript{\rm 2}Nanjing University of Science and Technology\\
    
    \{chenjiafu, wxing, csjk, chutianyi, ji\_by, cszhl, linhz, endywon\}@zju.edu.cn,\\ huangyiling@hotmail.com, hbchen@njust.edu.cn
}

\usepackage{bibentry}

\begin{document}

\maketitle

\begin{abstract}
	3D scene stylization refers to transform the appearance of a 3D scene to match a given style image, ensuring that images rendered from different viewpoints exhibit the same style as the given style image, while maintaining the 3D consistency of the stylized scene. Several existing methods have obtained impressive results in stylizing 3D scenes. However, the models proposed by these methods need to be re-trained when applied to a new scene. In other words, their models are coupled with a specific scene and cannot adapt to arbitrary other scenes. To address this issue, we propose a novel 3D scene stylization framework to transfer an arbitrary style to an arbitrary scene, without any style-related or scene-related re-training. Concretely, we first map the appearance of the 3D scene into a 2D style pattern space, which realizes complete disentanglement of the geometry and appearance of the 3D scene and makes our model be generalized to arbitrary 3D scenes. Then we stylize the appearance of the 3D scene in the 2D style pattern space via a prompt-based 2D stylization algorithm.
	Experimental results demonstrate that our proposed framework is superior to SOTA methods in both visual quality and generalization.
\end{abstract}

\section{Introduction}
3D scene stylization is an important editing task in vision and graphics, which facilitates the creation of new artistic scenes. Given a 3D scene and a style image, 3D scene stylization models can generate stylized images of the scene from arbitrary novel views. Naively applying approaches designed for image/video stylization to 3D scenes often leads to inconsistent results due to the lack of 3D information. To handle the inconsistency problem, several methods~\cite{huang2021learning,hollein2022stylemesh,cao2020psnet,kopanas2021point} have explored 3D scene stylization based on explicit representations ({\em e.g.}, meshes, voxels and point clouds). However, their discrete representation of scenes will lead to a loss of precision in geometry.

Recently, Neural Radiance Field (NeRF)~\cite{mildenhall2020nerf} proposes to use neural networks for continuous scene modelling. Due to its excellent performance in reconstructing both geometry and appearance, Stylizing-3D-scene~\cite{chiang2022stylizing} first introduces NeRF for 3D scene stylization. It fixes the geometry of the scene and changes only the appearance by using a hypernetwork to predict parameters for calculating artistic appearance. To further improve the visual quality of stylization, several approaches ~\cite{huang2022stylizednerf,zhang2022arf,nguyen2022snerf,fan2022unified,liu2023stylerf} have been developed. StylizedNeRF~\cite{huang2022stylizednerf} and SNeRF~\cite{nguyen2022snerf} are proposed to mutually optimize image stylization module and scene appearance representation to fuse the stylization ability of 2D stylization network with the 3D consistency provided by NeRF.  ARF~\cite{zhang2022arf} minimizes the distance between each feature vector in an image rendered from NeRF and its nearest neighbor feature vector in the given style image. INS~\cite{fan2022unified} simultaneously changes both the geometry and appearance to enable a more flexible stylization by stylizing shape tweaks on the scene surface. StyleRF~\cite{liu2023stylerf} transforms the grid features of the scene according to the reference style. However, despite valuable efforts, these methods still entangle geometry and appearance to some extent, necessitating re-training for a new scene.

In this work, we propose the {\bf first} framework for arbitrary 3D scene stylization, {\em which can not only transfer arbitrary styles but also stylize arbitrary 3D scenes with only one stylization model}. The key insight is a Prompt-based Neural Style Mapping (PNeSM) which disentangles the geometry and appearance of a 3D scene by mapping the appearance into a 2D style pattern space and then stylizes the appearance in the 2D style pattern space via a prompt-based 2D stylization algorithm. 

3D scene disentanglement consists of two main parts: UV mapping and appearance mapping.
Inspired by~\cite{xiang2021neutex}, {\em the UV mapping} trains a UV mapping network to project the 3D real-world coordinates into a 2D (UV) style pattern space. Different from~\cite{xiang2021neutex}, we use voxel-grid representation instead of MLP for fast training and use a stylization network rather than manual operation to change the appearance. Thanks to the complete separation of geometry and appearance, the stylization can be conducted in a unified style pattern space. {\em The appearance mapping} reconstructs the original appearance of the scene, which maps the projected style pattern coordinate to the radiance color through an MLP. 

3D scene stylization is realized via prompt-based stylization mapping. {\em The prompt-based stylization mapping} stylizes the appearance of the scene in the 2D style pattern space. Given arbitrary style images, a powerful pre-trained 2D stylization network ({\em e.g.}, SANet~\cite{park2019arbitrary}) can generate their corresponding 2D style patterns. However, directly using these 2D style patterns to stylize the appearance of 3D scenes would easily lead to disorganized results (as shown in Fig.~\ref{fig:prompt}). This is because they are generated and applied to the appearance of 3D scenes without taking the geometry information into consideration. To address this problem, we integrate a visual prompt to the feature maps of the bottleneck layer of the pre-trained 2D stylization network, and it is the only tensor we need to train in the stylization stage. 
When trained on a single 3D scene, the prompt can be treated as a scene-related adaptor, adapting the 2D style patterns to be aware of the specific geometry information of that scene. When trained on multiple 3D scenes, it can be treated as a scene-agnostic adaptor, adapting the 2D style patterns to be aware of the universal geometry information which is tolerant of diverse geometric variations. In our experiments, we find the prompt can generalize well to unseen scenes by learning on just few-shot ({\em e.g.}, 3) scenes. The scene-agnostic prompt thereby enables our framework to achieve arbitrary 3D scene stylization.

We have conducted comprehensive experiments to demonstrate the effectiveness and superiority of our proposed method. Experimental results demonstrate that our method not only achieves high-quality 3D scene stylization, but also generalizes well to unseen styles and unseen scenes.

Overall, the contributions can be summarized as follows:
\begin{itemize}
	\setlength{\itemsep}{0pt}
	\setlength{\parsep}{0pt}
	\setlength{\partopsep}{0pt}
	\setlength{\parskip}{0pt}
	\item We propose a novel 3D scene stylization framework, {\em i.e.}, PNeSM, which realizes complete disentanglement of the geometry and appearance of 3D scenes by mapping the appearance of 3D scenes into a 2D style pattern space. The stylization of 3D scenes is carried out in an independent and unified 2D style pattern space, which allows our framework to be generalized to any 3D scene. For each new 3D scene, there is no need to train a separate stylized model.
	
	\item To the best of our knowledge, we are the first to explore the use of prompt learning to adapt the pre-trained 2D stylization network for 3D scene stylization, which provides a simple yet effective way to improve the quality of stylized 3D scenes.
	
	\item Extensive experiments on different datasets are conducted to demonstrate the effectiveness and superiority of our method in visual quality when generalizing to new 3D scenes and new styles.
\end{itemize}  

\section{Related Work}

\subsection{Image/Video Style Transfer}
Image style transfer aims to create new artworks from real-world photos by using style information from real artworks.~\cite{gatys2016image} proposed an optimization-based style transfer method. However, the iterative optimization process is prohibitively slow. Motivated by this, several approaches~\cite{johnson2016perceptual,li2016precomputed,ulyanov2016texture,huang2017arbitrary,li2017universal,park2019arbitrary,liu2021adaattn} have been developed based on feedforward networks. With such rapid progress, satisfying artistic image can be easily generated.

Video style transfer~\cite{gao2018reconet,chen2017coherent,deng2021arbitrary,wang2020consistent} takes on the challenge of maintaining the consistency between adjacent frames in the stylized video, eliminating flickering effects. This is achieved by introducing optical flow or aligning intermediate feature to constrain nearby video frames.

Since both image and video style transfer can only stylize 2D images, lacking the knowledge of 3D scene, simply applying them to 3D scene stylization often leads to inconsistency between different views.

\begin{figure*}[t]
	\centering
	\includegraphics[width=\linewidth]{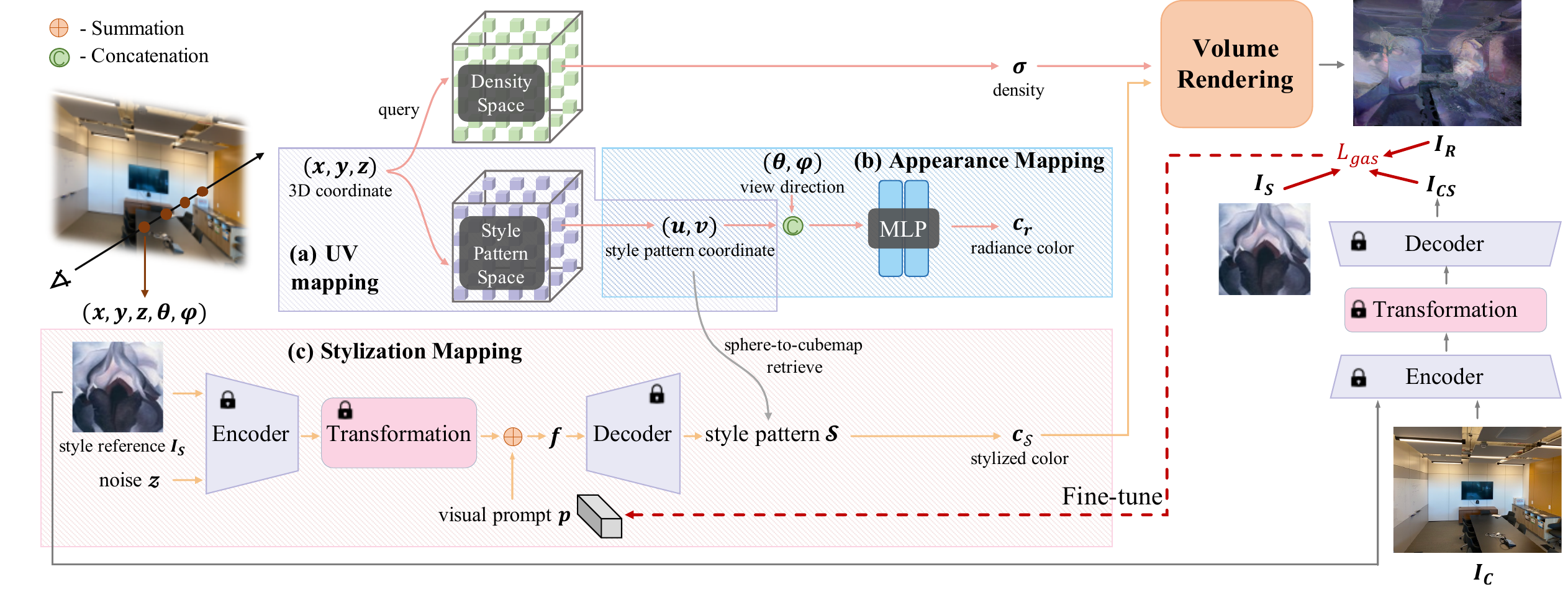}
	\caption{An overview of our method. (a) UV mapping is designed to support the complete disentanglement of geometry and appearance, which maps a 3D coordinate to a style pattern coordinate. (b) To reconstruct the original appearance, we use appearance mapping to map the style pattern coordinate along with the view direction to the radiance color $\mathbf{c_r}$. (c) A pre-trained image stylization network integrated with a visual prompt is used for stylization mapping, stylizing the appearance of the scene in the 2D style pattern space.}
	\label{fig:network}
\end{figure*}

\subsection{3D Scene Style Transfer}
3D scene style transfer requires that the images rendered from arbitrary viewpoints of the stylized scene match the style reference. Several approaches~\cite{huang2021learning,hollein2022stylemesh,cao2020psnet,kopanas2021point,mu20223d} have been developed using explicit 3D models ({\em e.g.}, meshes, voxels and point clouds). For example, \cite{huang2021learning} modulates scene features in point cloud with the given style image. 
However, these methods are limited by their quality of geometry reconstruction.

To offer a more faithful representation of scenes, some researchers turn to performing style transfer on radiance field~\cite{chiang2022stylizing,huang2022stylizednerf,zhang2022arf,nguyen2022snerf,fan2022unified,liu2023stylerf,chen2023testnerf}. Stylizing-3D-Scene~\cite{chiang2022stylizing} is the first to introduce NeRF to 3D scene style transfer, conducting patch-based optimization on content and style losses. StylizedNeRF~\cite{huang2022stylizednerf} and SNeRF~\cite{nguyen2022snerf} respectively propose mutual learning and alternate training strategy to effectively reduce GPU memory requirements. ARF~\cite{zhang2022arf} explores improving style details on stylized renderings with single style image and proposes a deferred back-propagation strategy to directly optimize on full-resolution images. INS~\cite{fan2022unified} proposes a method to interpolate between different styles in its pre-defined set and generates renderings by the new mixed styles. StyleRF~\cite{liu2023stylerf} conducts 3D scene stylization within the feature space of a radiance field and designs sampling-invariant content transformation to maintain multiview consistency. All of these methods require re-training a stylized model for every unseen scene. In this paper, our method focusses on not only transferring arbitrary styles, but also stylizing arbitrary 3D scenes with only one stylized model.

\subsection{Prompt Learning}
Prompt learning has emerged as a prominent technique in natural language process (NLP), with the hope to adapt pre-trained large language models, which are frozen, to downstream tasks by reformulating their input text. Building on domain-specific knowledge, some works~\cite{brown2020language,cui2021template,petroni2019language} manually design text prompts, achieving impressive results in few-shot or even zero-shot settings. To further unleash prompt's power, recent works propose to treat the prompt as task-specific variable and optimize it via backpropagation, namely Prompt Tuning~\cite{lester2021power,li2021prefix,liu2021p,zhong2021factual}.

Inspired by the success of prompt learning in NLP, researchers begin experimenting with applying prompts to computer vision. \cite{zhou2022learning} and~\cite{zhou2022conditional} transform context words into a set of learnable vectors for downstream image recognition, introducing prompt learning to vision-language models. VPT~\cite{jia2022visual} and VP~\cite{bahng2022exploring} explore prompting with images. VPT prepends a set of tunable parameters to ViT~\cite{dosovitskiy2020vit} in each Transformer encoder layer, which outperforms full fine-tuning ViT in many cases and reduces per-task storage cost. VP directly adds prompt as perturbations to the image in pixel space. These approaches show that prompt learning has great potential in visual domain. Motivated by visual prompt, we introduce prompt learning in feature level to style transfer field, adapting the pre-trained 2D image style transfer network for 3D scene stylization.

\section{Proposed Method}

As illustrated in Fig.~\ref{fig:network}, our proposed {\em Prompt-based Neural Style Mapping (PNeSM)} consists of three main parts: (a) A UV mapping that projects the 3D real-world coordinates into a 2D (UV) style pattern space, disentangling appearance from geometry. (b) An appearance mapping that maps the UV style pattern coordinate to the radiance color, representing the original appearance of the scene. (c) A prompt-based stylization mapping that stylizes the appearance of the scene in the 2D style pattern space, obtaining the final stylized color. To train PNeSM, we exploit a two-stage training strategy: I) a disentanglement stage which jointly trains the UV mapping and appearance mapping to reconstruct the scene, and II) a stylization stage which only trains the prompt-based stylization mapping for scene stylization. 

In the following subsections, we first provide a thorough review of our scene representation, NeRF, as preliminary. Then, we introduce how to completely disentangle appearance from geometry at the disentanglement stage. Finally, we introduce how to achieve stylization on the appearance at the stylization stage. 

\subsection{Preliminary} \label{sec:background}
NeRF~\cite{mildenhall2020nerf} proposes to encode a 3D scene as a function, $f: (x, y, z, \theta, \phi) \rightarrow (\sigma, \mathbf{c_r})$, which maps a 3D coordinate $(x, y, z)$ and its view direction $(\theta, \phi)$ to a volume density $\sigma$ and a radiance color $\mathbf{c_r}$. 

During volume rendering, rays $\mathbf{r}$ casting from the camera pass through the pixel of captured images. The pixel color thus can be calculated by sampling N points between $t_n$ and $t_f$ (the near and far bound):
\begin{equation}
	\begin{gathered}
		\hat{C}(\mathbf{r})=
		\sum\limits_{i=1}^{N} T_i(1-\exp(-\sigma_i\delta_i))\mathbf{c_r}_i, \\ \text{where } T_i=\exp(-\sum\limits_{j=1}^{i-1} \sigma_j\delta_j),
	\end{gathered}
	\label{eq:render}
\end{equation}
where $\delta_i=t_{i+1}-t_i$ denotes the distance between adjacent samples.

Given training images, NeRF model is optimized by minimizing the $L_2$ distance between the observed pixel $C(\mathbf{r})$ and the rendered pixel $\hat{C}(\mathbf{r})$:
\begin{equation}
	L_{rec} =\sum\limits_{\mathbf{r} \in \mathcal{R}} \parallel \hat{C}(\mathbf{r})-C(\mathbf{r}) \parallel _2,
\end{equation}
where $ \mathcal{R}$ is a ray batch from training views.

\begin{figure}[t]
	\centering
	\includegraphics[width=0.9\linewidth]{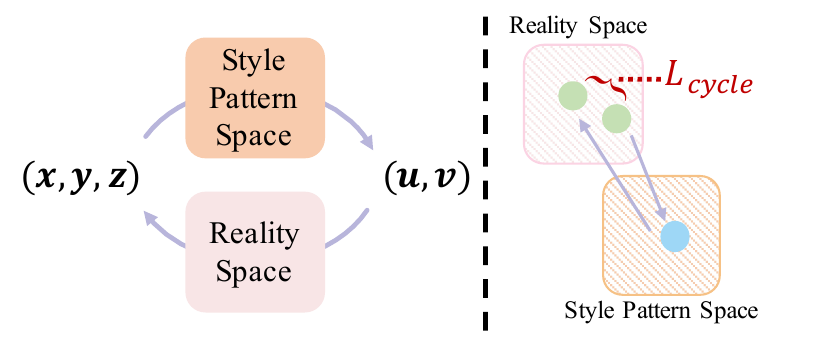}
	\caption{With the cycle loss $L_{cycle}$, we encourage the bijective mapping between real-world 3D coordinate and 2D style pattern coordinate.}
	\label{fig:cycle}
\end{figure}

\subsection{Appearance-Geometry Disentanglement} \label{sec:NeSM}
NeRF models radiance color ({\em i.e.}, scene appearance) using 3D coordinates and view directions as input. However, it entangles appearance and geometry in a ``black-box'' that cannot be edited. The target of 3D scene stylization is to stylize the appearance of the scene while retaining its geometry. Therefore, a critical desideratum is to disentangle appearance from geometry. Inspired by Neural Texture Mapping (NeuTex)~\cite{xiang2021neutex}, we add a UV mapping network to explicitly disentangle appearance from geometry by projecting real-world 3D coordinates into a 2D (UV) style pattern space during disentanglement stage. In this way, we can obtain the color of points on rays through their mapped style pattern coordinates, thereby enabling stylization of the scene's appearance in the unified 2D style pattern space. Each scene shares the same style pattern space. The disentanglement is achieved during conducting simple reconstruction training, where an appearance mapping (an MLP) is used to map the UV style pattern coordinate, along with view direction, to the radiance color $\mathbf{c_r}$, which represents the original appearance of the scene. After disentanglement, each UV style pattern coordinate in the style pattern space can pinpoint a specific point in the 2D style pattern via a sphere-to-cubemap retrieval operation (explained further in supp.). 
To speed up training, we use voxel-grid representation~\cite{sun2022direct} instead of MLP~\cite{xiang2021neutex} to model our UV mapping network and the modality of density.

Following~\cite{xiang2021neutex}, we also employ a cycle loss to ensure the rationality of the style pattern space, avoiding mapping multiple points in reality space to the same point in style pattern space. As shown in Fig.~\ref{fig:cycle}, we train another inverse mapping network to project the style pattern coordinate to reality space. In particular, for each ray, we focus more on whether the sample points that significantly contribute to the final pixel color maintain an accurate cycle mapping, reflecting the surface of the scene. From Eq.~(\ref{eq:render}), the contribution of each point is evident, so we consider it as the weight:
\begin{equation}
	w_i=T_i(1-\exp(-\sigma_i\delta_i)).
\end{equation}

The cycle mapping process and cycle loss are depicted in Fig.~\ref{fig:cycle} and defined as:
\begin{equation}
	(x, y, z)\rightarrow(u, v)\rightarrow(x^{\prime}, y^{\prime}, z^{\prime}),
\end{equation}
\begin{equation}
	L_{cycle}=\sum\limits_{i}w_i \|(x, y, z)-(x^{\prime}, y^{\prime}, z^{\prime})\|_2.
\end{equation}

The full loss function $L$ for scene reconstruction is:
\begin{equation}
	L=\lambda_{rec}L_{rec}+\lambda_{cycle}L_{cycle}.
\end{equation}

Note that though our appearance-geometry disentanglement of NeRF is based on NeuTex~\cite{xiang2021neutex}, there are two key differences: (1) The UV mapping network is formulated by voxel-grid representation~\cite{sun2022direct} instead of MLP~\cite{xiang2021neutex}, which can greatly speed up training while maintaining the reconstruction quality. (2) We add a new prompt-based stylization mapping to stylize the appearance of the scene in the 2D style pattern space, which is aware of the geometry information of the scene and can stylize the appearance more harmoniously.

\begin{figure*}[t]
	\centering
	\includegraphics[width=0.938\linewidth]{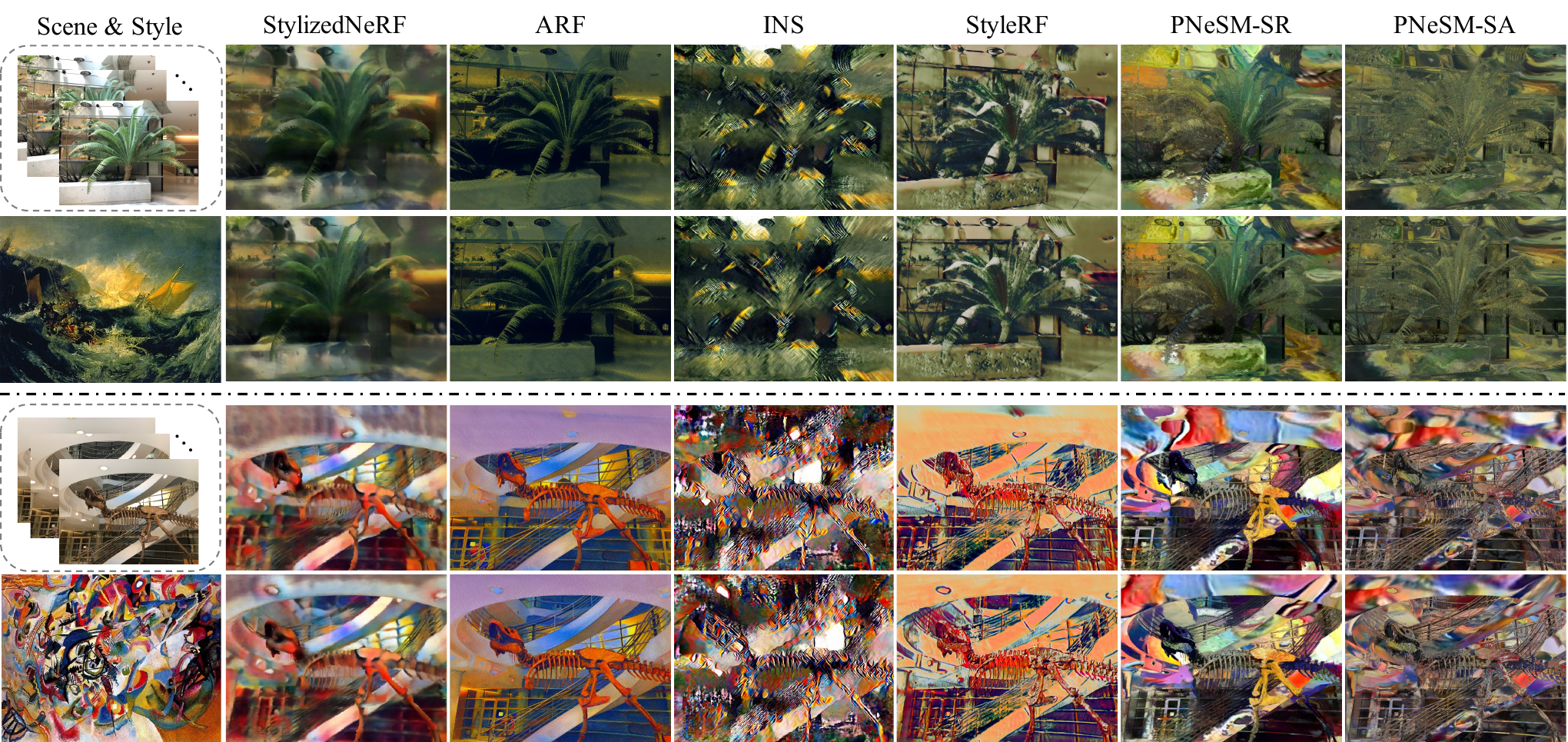}
	\caption{Qualitative comparisons on LLFF dataset. We compare our method to StylizedNeRF~\cite{huang2022stylizednerf}, ARF~\cite{zhang2022arf}, INS~\cite{fan2022unified} and StyleRF~\cite{liu2023stylerf}. Our method stylizes scenes with clear geometry and competitive stylization quality.} 
	\label{fig:llff}
\end{figure*}

\subsection{Prompt-based Appearance Stylization} \label{sec:prompt}
After the disentanglement stage, we can intuitively stylize the appearance of the scene in the unified 2D style pattern space via stylization mapping. We first generate a new 2D style pattern $\mathcal{S}$ given a reference style image $I_S$. Next, we locate each UV coordinate in the style pattern space to the corresponding pixel in the 2D style pattern via sphere-to-cubemap retrieval. Specifically, for a UV coordinate $(u, v)$, we retrieve the pixel $\mathbf{c_\mathcal{S}}$ in the 2D style pattern. $\mathbf{c_\mathcal{S}}$ is the stylized color for $(u, v)$, signifying the newly stylized appearance of the scene. 

As demonstrated in~\cite{gatys2016image,li2017universal}, pre-trained 2D stylization approaches are effective in extracting the texture information from style images. To learn only style patterns and remove contents in the style image, a noise image ${\boldsymbol z}$ is utilized as the content input.
Subsequently, the style patterns are employed to change the appearance of the scene within the style pattern space. However, due to the lack of consideration on scene's geometry, directly using the style patterns explained above would easily lead to disorganized results, as will be demonstrated later in Fig.~\ref{fig:prompt}. A satisfactory stylized scene should not only exhibit pleasant style patterns in the appearance, but also harmoniously fuse the style patterns with the scene geometry inherently. Therefore, the geometric awareness is important for 3D scene stylization and must be properly considered.


In order to integrate geometric information into the generated style patterns, we can fine-tune the decoder of a pre-trained 2D stylization network under the supervision of a {\em geometry-aware stylization loss} $L_{gas}$:
\begin{equation}
	\begin{aligned}
		L_{gas} =\sum\limits \parallel I_{CS}-I_R(\boldsymbol \theta) \parallel _2 \\ 
		+ \lambda_{style}\sum_{i}\|\mu(\phi_i(I_{S}))-\mu(\phi_i(I_R(\boldsymbol \theta)))\|_2 \\ 
		+ \lambda_{style}\sum_{i}\|s(\phi_i(I_{S}))-s(\phi_i(I_R(\boldsymbol \theta)))\|_2, \\
		{\boldsymbol \theta^*}=\mathop{\arg\min}\limits_{\boldsymbol \theta}L_{gas},
	\end{aligned}
	\label{eq:gas}
\end{equation}
where $I_R(\boldsymbol \theta)$ denotes the rendered image given the fine-tuned 2D stylization network ${\boldsymbol \theta}$ and $I_{CS}$ denotes the stylized image using training views as content inputs to pre-trained 2D stylization network. $\mu$ and $s$ are channel-wise mean and standard deviation, respectively. $\phi_i$ denotes a layer in VGG-19. The first term aligns stylized training views and rendered views from the scene, thus the style patterns generated by the 2D stylization network should be aware of the geometry information of the scene. The last two terms calculate the style loss between rendered images and the style reference, in the manner typically employed in image stylization methods.

However, fine-tuning the decoder of a pre-trained 2D stylization network is cumbersome and time-consuming, making it inflexible in practical. To alleviate this problem, 
we introduce prompt learning for fast and flexible adaptation. To be specific, we add a visual prompt ${\boldsymbol p}$ to the output feature maps from the style transformation module of the 2D stylization network. The visual prompt is treated as an extra and independent learnable component implicitly representing geometry information of scenes. We train the visual prompt using $L_{gas}$ to generate more harmonious style pattern for scenes during stylization. All parameters of image stylization network are frozen, and the visual prompt is the only parameter requires training at the stylization stage. This means that ${\boldsymbol \theta}$ in Eq.~\ref{eq:gas} corresponds to ${\boldsymbol p}$ in our method. Note that our method is not limited to a specific 2D stylization network and the ability to transfer arbitrary styles is inherently embedded within the arbitrary image stylization network. The visual prompt is plug-and-play and can be easily integrated into existing image style transfer methods. 


		%

\begin{figure*}[t]
	\centering
	\includegraphics[width=0.91\linewidth]{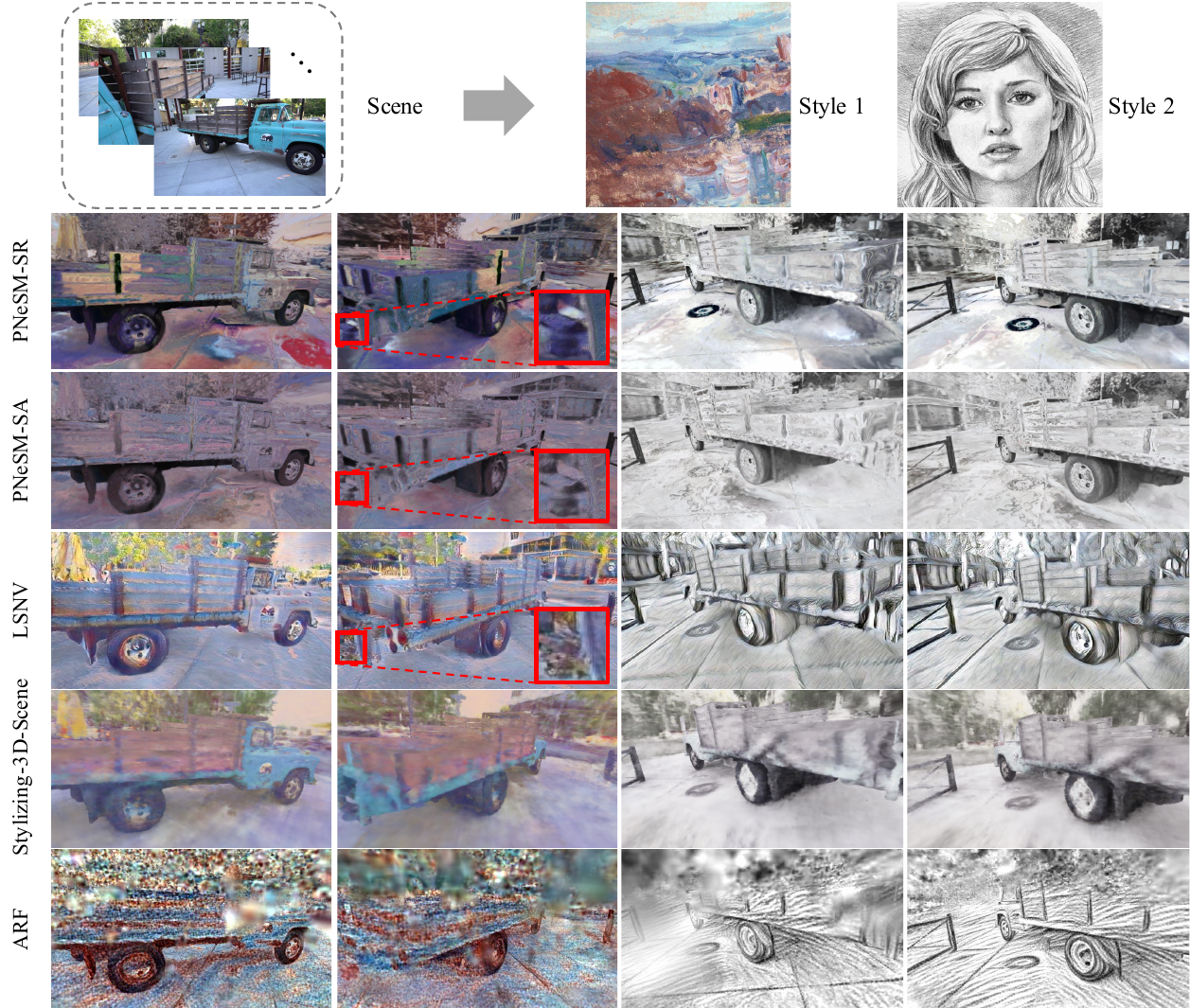}
	\caption{Qualitative comparisons on Tanks and Temples dataset. We compare our method to LSNV~\cite{huang2021learning}, Stylizing-3D-Scene~\cite{chiang2022stylizing} and ARF~\cite{zhang2022arf}. Stylized scenes generated by our method contain both precise geometry and pleasant stylization.}
	\label{fig:tnt}
\end{figure*}

\section{Experiments}

\subsection{Implementation Details}
We implement our model during the disentanglement stage on top of DVGO~\cite{sun2022direct}, where we replace the feature grid as described in~\cite{sun2022direct} with a grid designed for style pattern space. Following~\cite{sun2022direct}, we use the Adam optimizer with a learning rate of 0.1 for all voxels and 0.001 for MLP. $\lambda_{rec}$ and $\lambda_{cycle}$ are set to $1$. During stylization, we adopt SANet~\cite{park2019arbitrary} as the image style transfer network. 
The visual prompt is trained for 5k iterations using an Adam optimizer with a learning rate of 0.1. $\lambda_{style}$ is set to $0.1$. We use $\rm relu1\_1$, $\rm relu2\_1$, $\rm relu3\_1$, and $\rm relu4\_1$ layers in VGG-19 to calculate loss in Eq.~\ref{eq:gas}. Appearance-geometry disentanglement is scene-related, while prompt-based appearance stylization is scene-agnostic. For arbitrary test scenes, their appearance and geometry should be disentangled first, and then the stylization can be conducted by the stylization mapping module trained on training scenes. All experiments are performed on a single NVIDIA RTX A6000 (48G) GPU.

{\bf Datasets.}  Following previous image stylization methods, we take WikiArt~\cite{karayev2013recognizing} as the style dataset. We conduct extensive experiments on real-world scenes, forward-facing LLFF~\cite{mildenhall2019local} and $360^{\circ}$ unbounded Tanks and Temples dataset~\cite{knapitsch2017tanks}. The training sets of LLFF dataset are {\em Room, Horns, Leaves, Flower, Orchids}, and we use {\em Fern, Trex} for evaluation. On Tanks and Temples dataset, we use {\em Playground, Horse, Francis} for training, and evaluate on {\em Truck}.

{\bf Baselines.} On LLFF dataset, we compare our method to StylizedNeRF~\cite{huang2022stylizednerf}, ARF~\cite{zhang2022arf}, INS~\cite{fan2022unified} and StyleRF~\cite{liu2023stylerf}. On Tanks and Temples Dataset, we compare our method to LSNV~\cite{huang2021learning}, Stylizing-3D-Scene~\cite{chiang2022stylizing} and ARF~\cite{zhang2022arf}. For all these methods, we use their released codes and pre-trained models. Among them, LSNV is based on point cloud scene representation, while others are based on NeRF. We do not conduct comparison on image/video style transfer methods, which are less competitive than 3D scene stylization approaches proven in previous works~\cite{huang2021learning,huang2022stylizednerf,nguyen2022snerf,chiang2022stylizing}.

\subsection{Qualitative Results}
We experiment with both scene-related (PNeSM-SR) and scene-agnostic (PNeSM-SA) visual prompt on our method.

{\bf LLFF.} 
In Fig.~\ref{fig:llff}, we show qualitative comparisons 
on LLFF dataset. We observe that StylizedNeRF~\cite{huang2022stylizednerf} degrades the scene in clarity, which might be caused by introducing spatial consistency to 2D stylization network and training style module for NeRF with the supervision of fine-tuned 2D stylization results. ARF~\cite{zhang2022arf} sometimes produces plain results in the aspect of color tone ({\em e.g.} 3rd and 4th rows). INS~\cite{fan2022unified} disrupts the geometry of scenes, yielding poor-quality stylizations. StyleRF~\cite{liu2023stylerf} shows low similarity between stylized scenes and style images. In contrast, our method can not only maintain clear geometry, but also change the appearance of the scene resembling the reference style. Our method shows better stylization quality in terms of style transformation. (Please refer to the quantitative comparison on {\em style loss} in supp.)

{\bf Tanks and Temples.} 
In Fig.~\ref{fig:tnt}, we qualitatively compare our results with baselines on Tanks and Temples dataset. LSNV~\cite{huang2021learning} reconstructs the scene with point cloud, whose geometry is not precise and further damages the stylization result. Stylizing-3D-Scene~\cite{chiang2022stylizing} calculates Gram matrix loss~\cite{gatys2016image} on sub-sampled patches to achieve stylization. Due to the limited receptive field, the stylized results are blurry and the stylization quality is poor. ARF~\cite{zhang2022arf} contains geometry artifacts for Tanks and Temples dataset on their implementation based on Plenoxel~\cite{fridovich2022plenoxels}, which is also mentioned in their {\em Limitations}. Therefore, the quality of stylized renderings is also affected. Our approach generates both precise geometry and stylization following the artistic style of the style reference.


\subsection{Quantitative Results}

Following the measurement in LSNV, we use a warped LPIPS metric~\cite{zhang2018unreasonable} to measure the consistency across different views. We utilize FlowNetS~\cite{dosovitskiy2015flownet} to compute the optical flow from a ground truth image $I_x$ to another $I_y$. Subsequently, a warped mask $M$ is generated based on the optical flow. Finally, we warp the corresponding stylized images $\hat{I}_x$ to $\hat{I}_y$ and calculate their distance along with $M$. The distance score is formulated as:
\begin{equation}
	E(\hat{I}_x,\hat{I}_y)=\mathit{LPIPS}(M \odot \mathit{Warp}(\hat{I}_x,\hat{I}_y)),
\end{equation}
where $\odot$ denotes element-wise multiplication.

We compare our method with baselines on LLFF dataset, reporting average warped distance score on 5 style references. We randomly choose 20 frame pairs $(\hat{I}_t,\hat{I}_{t+1})$ and $(\hat{I}_t,\hat{I}_{t+7})$ from each scene for short-range and long-range consistency respectively.


\subsection{Ablation Study}
\begin{figure}[t]
	\centering
	\setlength{\tabcolsep}{0.03cm}
	\begin{tabular}{ccc}
		\footnotesize Scene \& Style & \footnotesize (a) &\footnotesize (b)
		\\
		\includegraphics[width=0.29\linewidth]{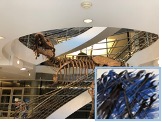}&
		\includegraphics[width=0.29\linewidth]{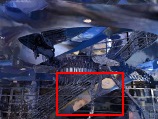}&
		\includegraphics[width=0.29\linewidth]{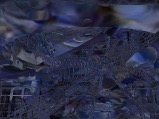}
		\vspace{-1mm}
		\\
		\includegraphics[width=0.29\linewidth]{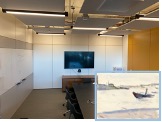}&
		\includegraphics[width=0.29\linewidth]{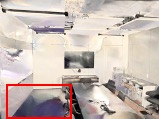}&
		\includegraphics[width=0.29\linewidth]{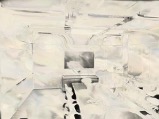}
	\end{tabular}
	\caption{Ablation study on direct image stylization on reconstruction appearance. (a) The results of using reconstruction appearance cubemap as content input for image stylization. (b) The results of our method (using a noise image as content input and add a visual prompt in the bottleneck of image stylization network.)} 
	\label{fig:noise}
\end{figure}

{\bf Direct Image stylization on reconstruction appearance.} We inverse sphere-to-cubemap retrieval to extract a cubemap showing the reconstruction appearance of the scene and use the cubemap as content input of the image stylization network instead of noise ${\boldsymbol z}$. We report the experimental results in Fig.~\ref{fig:noise}, where we observe there is abrupt color in some areas impairing the stylization quality. We suggest that this is because the appearance is not uniformly distributed on the style pattern space. Thus, a small region in the cubamap might represent a wide area of the scene appearance, magnifying abrupt color in the appearance.

\begin{figure}[t]
	\centering
	\includegraphics[width=0.9\linewidth]{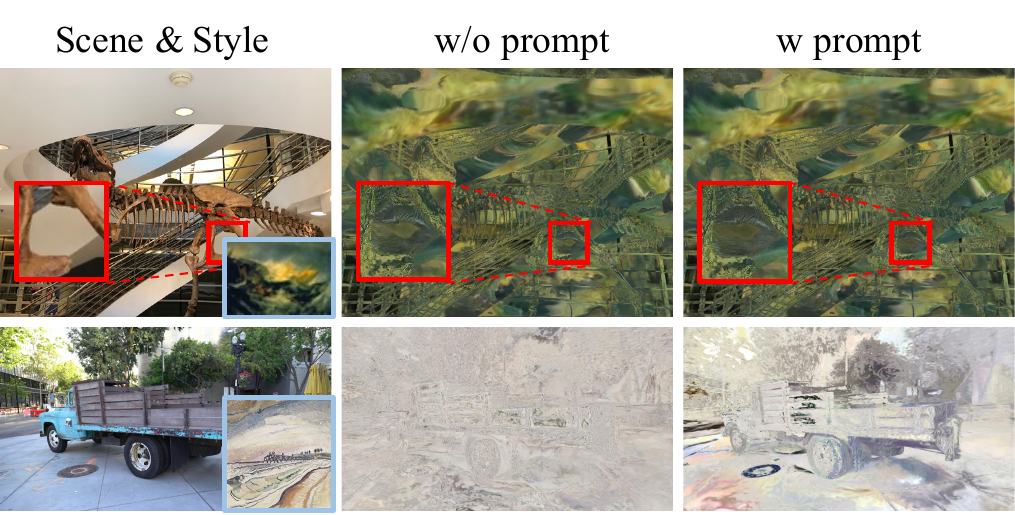}
	\caption{Ablation study for the visual prompt. The visual prompt alleviates the disorganized results and improve the visual quality.}
	\label{fig:prompt}
\end{figure}

\begin{table}[t]
	\small
	\centering
	\tabcolsep=3pt
	\begin{tabular}{c|ccccc}
		\hline
		Methods & StylizedNeRF & ARF & INS & StyleRF & PNeSM\\
		\hline
		Short-range & 0.0229 & 0.0125 & 0.0208 & 0.0235 & 0.0116\\
		Long-range & 0.0627 & 0.0353 & 0.0439 & 0.0531 & 0.0351\\
		\hline
	\end{tabular}
	\caption{Short-range and Long-range consistency comparison. The lower the better.}
	\label{table:consistency}
\end{table}

\noindent{\bf With and without visual prompt.} To investigate the effect of introducing a visual prompt, we evaluate the performance when it is removed. The result in Fig.~\ref{fig:prompt} shows that directly using image style transfer network to generate style patterns can realize stylization, but the results are obviously disorganized without considering geometry information. It demonstrates that the visual prompt helps to adapt the style patterns to be aware of geometry information, thus the appearance of the scene can be stylized more harmoniously.


\section{Conclusion}
In this paper, we present a Prompt-based Neural Style Mapping (PNeSM) to transfer arbitrary styles to arbitrary 3D scenes. We take advantage of the powerful reconstruction capability of NeRF and completely disentangle appearance and geometry by mapping the appearance into a 2D style pattern space. By fusing the ability of texture information extraction in pre-trained 2D stylization network and effectiveness of prompt learning for fine-tuning, we achieve pleasant 3D scene stylization by stylizing the appearance of the scene in the 2D style pattern space. Extensive experimental results demonstrate the effectiveness and superiority of our method.

\section*{Acknowledgements}
This work was supported in part by Zhejiang Province Program (2022C01222, 2023C03199, 2023C03201, 2019007, 2021009), the National Program of China (62172365, 2021YFF0900604, 19ZDA197), Ningbo Program(2022Z167), and MOE Frontier Science Center for Brain Science \& Brain-Machine Integration (Zhejiang University).

\bibliography{aaai24}

\end{document}